\documentclass[11pt]{article}

\usepackage{acl}

\usepackage{times}
\usepackage{latexsym}
\usepackage{graphicx}
\usepackage[T1]{fontenc}

\usepackage[utf8]{inputenc}
\usepackage{microtype}
\usepackage{booktabs}
\usepackage{multirow}
\usepackage{inconsolata}

\usepackage{array}
\usepackage{longtable}
\usepackage{colortab}
\usepackage{colortbl}
\usepackage{arydshln}

\newcommand{\footlink}[1]{\footnote{\url{#1}}}

\newcommand{\tabH}{\rule{0pt}{2.1ex}}

\title{Improving Sentence Embeddings with Automatic Generation of\\Training Data Using Few-shot Examples}

\author{Soma Sato\hspace{2em} Hayato Tsukagoshi \hspace{2em} Ryohei Sasano \hspace{2em} Koichi Takeda \\
  Graduate School of Informatics, Nagoya University\\
 \texttt{\{sato.soma.y7,tsukagoshi.hayato.r2\}@s.mail.nagoya-u.ac.jp}\\
 \texttt{\{sasano,takedasu\}@i.nagoya-u.ac.jp}}

\begin{document}
\maketitle

\begin{abstract}
Decoder-based large language models (LLMs) have shown high performance on many tasks in natural language processing. 
This is also true for sentence embedding learning, where a decoder-based model, PromptEOL, has achieved the best performance on semantic textual similarity (STS) tasks. 
However, PromptEOL requires a manually annotated natural language inference (NLI) dataset for fine-tuning.
We aim to improve sentence embeddings without using large manually annotated datasets by automatically generating an NLI dataset with an LLM and using it for fine-tuning of PromptEOL.
To achieve this, we explore methods of data generation suitable for sentence embedding learning in this study.
Specifically, we will focus on automatic dataset generation through few-shot learning and explore the appropriate methods to leverage few-shot examples.
Experimental results on the STS tasks demonstrate that our approach outperforms existing models in settings without large manually annotated datasets.\footnote{Our code is available at \url{https://github.com/lamsoma/Auto_NLI}}
\end{abstract}

\section{Introduction}
Sentence embeddings are widely studied as they can be used for many tasks such as text search, entailment recognition, and information extraction~\cite{SentenceBERT,DefSent,SimCSE,PromptBERT,ST5}. 
Among these, methods based on decoder-based large language models (LLMs) have shown high performance in recent years.  
For example, SGPT~\cite{sgpt}, which uses decoder-based LLMs to generate embeddings, and PromptEOL~\cite{PromptEOL}, which generates sentence embeddings using a prompt-based method focusing on a single word, have been proposed. 
PromptEOL achieves the highest performance in STS in a setting using manually annotated data. 
However, when not using manually annotated NLI datasets, its performance is much lower.

Since the advent of high-performance decoder-based LLMs like GPT-4,\footlink{https://openai.com/index/gpt-4-research/} many efforts have been made to use data generated by decoder-based LLMs as a substitute for training data in various tasks, and their effectiveness has been reported~\cite{GeneratingData,ZeroGen,ProGen}.
Similarly, for sentence embedding learning, there are approaches such as GenSE~\cite{GenSE}, which automatically generates datasets using LLMs to augment sentence embedding datasets, and STS-Dino~\cite{Dino-STS}, which is an automatically generated dataset for training sentence embedding models using LLMs.
However, there has not been sufficient investigation on how to generate data using LLMs for sentence embedding learning.
It is known that when generating datasets automatically via few-shot learning, the generated datasets are heavily dependent on the few-shot examples~\cite{Calibrate}, and if all the data is generated by using the same few-shot examples, the diversity of the generated datasets may be limited.

In this study, we explore how few-shot examples should be leveraged to automatically generate training data to obtain better sentence embeddings in a framework where NLI datasets generated by an LLM are used for fine-tuning of PromptEOL.
Specifically, we examine how the quality of the final sentence embeddings varies when the number of few-shot examples used to generate training data is varied or when multiple sets of few-shot examples are used, and we reveal the optimal way to leverage few-shot examples.
Our contributions are two-fold.
First, we explored the optimal ways to leverage few-shot examples when using LLMs to generate NLI datasets for sentence embedding learning.
Second, we achieved the highest score in the STS tasks in a setting without large manually annotated datasets.
\section{Related Work}

This section introduces PromptRoBERTa~\cite{PromptBERT} and PrompEOL~\cite{PromptEOL}, which successfully generate high-performance sentence embeddings by devising prompts. 

\paragraph{PromptRoBERTa}
PromptRoBERTa introduces a new contrastive learning method to improve sentence embedding performance of RoBERTa. 
Specifically, it takes a sentence like ``I have a dog.'' as input and transforms it using templates as follows:  ``\texttt{This sentence: "I have a dog." means [MASK].}'' and ``\texttt{The sentence: "I have a dog." means [MASK].}'' . 
By using the embeddings of the ``\texttt{[MASK]}'', it can represent the same sentence from diverse perspectives using different templates, resulting in reasonable positive pairs of sentence embeddings.
By learning to bring these positive pairs of sentence embeddings closer together, PromptRoBERTa significantly reduces the performance gap between supervised and unsupervised settings, achieving better sentence embedding performance compared to traditional methods.

\paragraph{PromptEOL}
PromptEOL introduces a constraint called the ``one-word limitation'' and inputs the target sentence into LLMs along with a template.
For example, to obtain the embedding of the sentence ``I have a dog.'', it inputs the prompt ``\texttt{This sentence: "I have a dog." means in one word: "}'' into a decoder-based LLM.
The hidden vector after ``\texttt{in one word: "}'' is then used as the sentence embedding.
Since the decoder-based LLM is pretrained on the next-token prediction task, it can obtain a sentence embedding that captures the meaning of the whole sentence by using the prompt to predict a word that paraphrases the entire sentence.
Although PromptEOL demonstrates relatively high performance in an unsupervised setting, it can produce higher quality embeddings through supervised learning.
PromptEOL achieved the best performance in the STS tasks by fine-tuning the LLM via contrastive learning on NLI datasets similar to supervised SimCSE~\cite{SimCSE}.

\section{Automatic NLI Dataset Generation}

In this study, we explore how to automatically construct datasets for sentence embedding learning using LLMs. 
In this section, we explain the process of generating NLI datasets with LLMs.

\subsection{Existing NLI Datasets}
NLI datasets are widely used in various sentence embedding models, including SimCSE~\cite{SimCSE} and PromptEOL~\cite{PromptEOL}.
They contain sentence pairs comprising a premise and a hypothesis, which is labeled with either ``entailment,'' ``neutral,'' or ``contradiction.''
The prominent NLI datasets include the Stanford NLI (SNLI) corpus~\cite{snli:emnlp2015}, the Multi-Genre NLI (MNLI) corpus~\cite{N18-1101}, and the Cross-Lingual NLI (XNLI) corpus~\cite{xnli}, which contain approximately 579,000, 433,000, and 112,500 sentence pairs, respectively. 
Following \citet{PromptEOL}, we use a dataset that combines the SNLI and MNLI corpora, and refer to it as the manual NLI dataset.

\subsection{Automatic Generation Procedure}
To automatically build NLI datasets, we generate hypothesis sentences from premise sentences.
Specifically, we replace \texttt{[premise]} in each of the following prompts with a premise sentence and then feed the prompt to the LLM.
\begin{description}
\setlength{\itemindent}{-0.6em}
\setlength{\leftskip}{-1.2em}
\item[Prompt for entailment] \texttt{\small \\
Write one sentence that is logically entailed by [premise] in the form of a statement beginning with "Answer: ". Answer: "}
\item[Prompt for contradiction] \texttt{\small \\
Write one sentence that logically contradicts [premise] in the form of a statement beginning with "Answer: ". Answer: "}
\end{description}
Next, we take the tokens generated between ``\texttt{Answer: \hspace{-0.4em}"}'' and the next ``\texttt{"}'' as the generated hypothesis sentence.

We further improve the quality of the generated hypothesis sentences by applying few-shot learning~\cite{NEURIPS2020_1457c0d6}. Specifically, we extract a few sentence pairs from the manual NLI dataset and add them as few-shot examples.
The number of examples is around 20 at most, which is a feasible amount even if created manually from scratch.

\section{Experiments}
We first evaluate automatically generated NLI datasets using NLI classifiers.
Next, we evaluate sentence embedding models fine-tuned with automatically generated NLI datasets.
We explore how to use few-shot examples specifically for sentence embedding learning.
After conducting these experiments, we compare the best-performing method from our exploration with existing methods.

\subsection{Evaluation of NLI Dataset}
\label{sec:nli-classification}

We evaluated the quality of the automatically generated NLI datasets using an NLI classifier.
This allows us to assess the quality of NLI datasets generated by LLMs automatically.

\paragraph{Generation Method}

In our method, we generate hypotheses according to premises as input.
Therefore, for the source premise sentences, we randomly extracted one million sentences from Wikipedia, following the unsupervised fine-tuning dataset of SimCSE~\cite{SimCSE}.
To reduce potential biases from the difference between the manual NLI datasets and sentences from Wikipedia,
we used sentences with token counts between 4 and 32 to approximate the distribution of the manual NLI dataset.
The frequency distribution of the token count is shown in Appendix~\ref{appendix:distribution}.
We used LLaMA-2-7B-Chat~\cite{llama2} as the LLM.

\begin{table}[t]
\small
  \centering
  \begin{tabular}{ccc}
    \hline
    \tabH Dataset & Entailment  &  Contradiction  \\
    \hline
    \tabH 0-shot & 0.348 & 0.901 \\
    1-shot & 0.627 & 0.830 \\
    5-shot & 0.883 & 0.941 \\
    20-shot & 0.944 & 0.949 \\
    \hline
    \tabH Manual NLI dataset & 0.929 & 0.941 \\
    \hline
\end{tabular}
\caption{The agreement ratio between the predicted and assigned labels of NLI datasets generated with zero-/few-shot learning and the manual NLI dataset
}
\label{table:data_type}
\end{table}

\paragraph{Evaluation Method}
We used DeBERTa~\cite{he2021deberta} trained on the MNLI corpus.\footlink{https://huggingface.co/microsoft/deberta-v2-xxlarge-mnli}
For each sentence pair in the generated dataset, we performed a three-way classification of entailment, neutral, or contradiction.
We then calculated the agreement ratio between the classification result and the assigned labels.
For the manual NLI dataset and a zero-shot generated dataset, we randomly selected 3,000 sentence pairs for both entailment and contradiction, totaling 6,000 pairs, and calculated the agreement ratios for these pairs.
For the few-shot generated datasets, to mitigate randomness due to the few-shot examples, we first created 10 sets of different examples for both entailment and contradiction.
Then, each set was given 1,000 different premise sentences to create pairs, resulting in 20,000 pairs for evaluation.

\paragraph{Experimental Results}
Table~\ref{table:data_type} lists the agreement ratios for each dataset.
The ratio improved as the number of few-shot examples increased.
In the 5-shot setting, the ratio of contradiction is comparable to that of the manual NLI dataset, and in the 20-shot setting, the ratio for both entailment and contradiction reached levels comparable to those of the manual NLI dataset.
These results suggest that the automatically generated NLI datasets with 5-shot or 20-shot learning were reasonably high quality.
We provide examples of datasets obtained with 0-shot and 20-shot learning in Appendix~\ref{appendix:NLI_examples}.

\subsection{Explore How to Use Few-shot Examples}
\label{sec:sts}

We evaluated sentence embedding models fine-tuned with the automatically generated NLI datasets using the STS tasks.\footnote{Evaluations were also conducted on downstream tasks of SentEval~\cite{SentEval}, but as reported in~\citet{PromptEOL}, the effectiveness of fine-tuning with the NLI dataset could not be confirmed.
We provide the results of downstream tasks in Appendix~\ref{appendix:transfer_tasks}.
}
The STS task is to evaluate whether a model could correctly estimate the semantic similarity of sentence pairs.
Specifically, we calculated the semantic similarity via the model and tested its closeness to a human evaluation.
Following previous studies~\cite{SentenceBERT,SimCSE,PromptEOL}, the sentence embedding quality was evaluated in terms of Spearman's rank correlation coefficient between the cosine similarity of sentence embeddings and the human ratings.


\paragraph{Experimental Setup}
We fine-tuned LLaMA-2-7B~\cite{llama2} with NLI datasets.
Following \citet{PromptEOL}, we used the same seven STS datasets for evaluation: STS 2012--2016~\cite{STS12,STS13,STS14,STS15,STS16}, STS-B~\cite{STS-B}, and SICK-R~\cite{Sick}.
To investigate the relationship between dataset size and performance, we trained our model with different numbers of examples.
The number of examples in the datasets is 4,000$\times 2^n$ (n = 0, 1, …, 6).
For fine-tuning with PromptEOL, we used NLI datasets generated with 0-shot, 1-shot, 5-shot, 20-shot, 1-shot$\times$5 (five combined 1-shot datasets), 5-shot$\times$4 (four combined 5-shot datasets) setups and the manual NLI dataset.
We used the same hyperparameters as PromptEOL~\cite{PromptEOL}, with a batch size of 256 during training, 10\% of the total steps for warm-up, and a learning rate of 5e-4.
During training, we calculated Spearman’s rank correlation coefficient on the STS-B development set every (number of data / 4000) step and used the model with the highest score for the final evaluation.
To minimize randomness from few-shot examples, we generated multiple NLI datasets: 10 for 1-shot, 5 for 1-shot$\times$5 and 5-shot, 4 for 5-shot$\times$4 and 20-shot, and 3 for zero-shot.
We report their average scores for the final evaluation.

\begin{figure}[t]
    \centering
    \includegraphics[width=1.0\columnwidth]{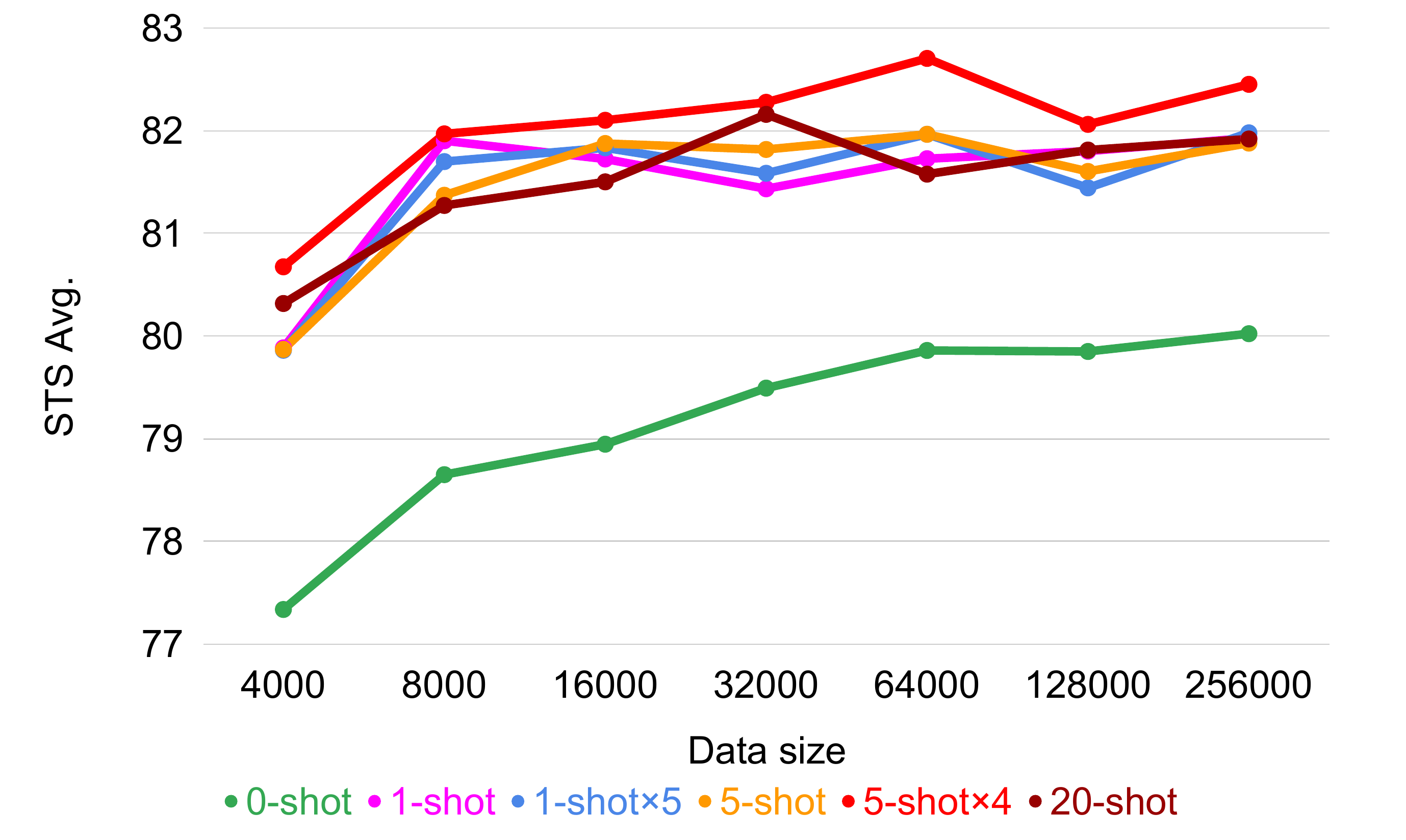}
    \caption{Performances of different few-shot settings}
    \label{fig:4.21}
\end{figure}

\paragraph{Experimental Results}
Figure~\ref{fig:4.21} shows the results. 
Comparing the zero-shot and few-shot results, the few-shot performances outperformed the zero-shot performance regardless of the amount of data size, thus confirming the effectiveness of few-shot learning.
Comparing 1-shot, 5-shot, and 20-shot, there there was no improvement in scores as the number of shots increased. 
This indicates that merely increasing the number of shots does not necessarily lead to better performance.
Although there was little performance difference between 5-shot and 1-shot$\times$5, 5-shot$\times$4 consistently outperformed 20-shot, regardless of data size.
According to Section~\ref{sec:nli-classification}, although the quality of the generated dataset with 1-shot learning is not sufficient, the generated dataset with 5-shot learning has sufficiently high quality.
This suggests that distributing few-shot examples can improve performance, but only when the data quality exceeds a certain threshold.
5-shot$\times$4 successfully introduces diversity while maintaining sufficient quality, and this balance between diversity and quality appears to be crucial for enhancing the effectiveness of sentence embeddings generated from NLI datasets.



\subsection{Comparison with Existing Methods}\label{sec:4.3}
To evaluate the performance of models trained on automatically generated NLI datasets, we compared the following five models:
1) PromptEOL without fine-tuning,
2) PromptEOL fine-tuned with the generated dataset using 0-shot learning,
3) PromptEOL fine-tuned with the generated dataset using 5-shot$\times$4 learning,
4) PromptEOL fine-tuned with the manual NLI dataset,
5) Unsupervised PromptRoBERTa~\cite{PromptBERT}, which achieved the highest performance without using manually annotated large-scale datasets.
For unsupervised PromptRoBERTa, we used the premise sentences to automatically generate NLI datasets, which are used for training.
For PromptRoBERTa and experiments using manually annotated datasets, we conducted experiments three times with different random seeds, and we reported their average scores as the final score.
Other experimental settings and evaluation methods were the same as in Section~\ref{sec:sts}.

\begin{figure}[t]
    \centering
    \includegraphics[width=1.0\columnwidth]{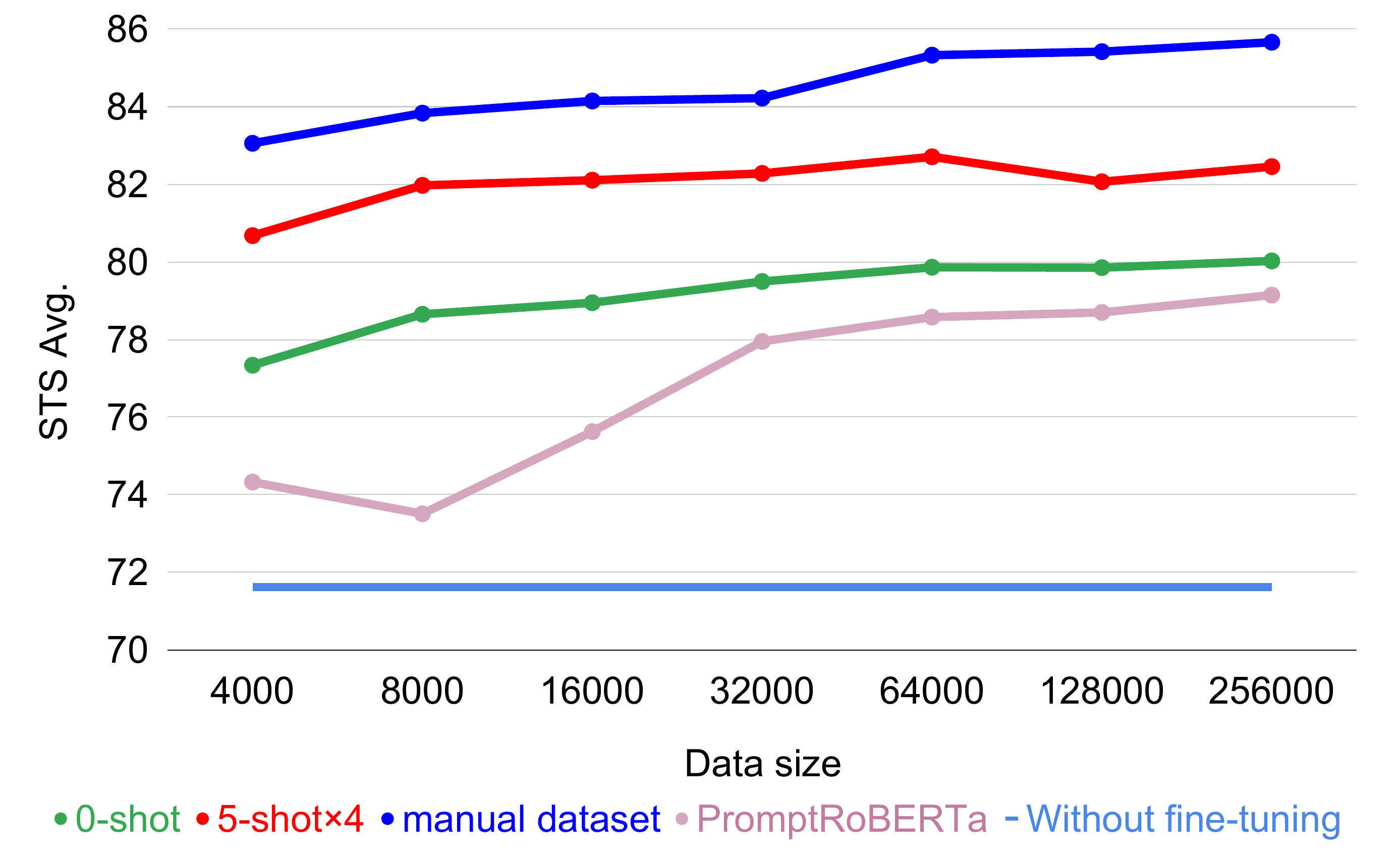}
    \caption{Performances of models fine-tuned with the automatically generated datasets and existing methods}
    \label{fig:4.22}
\end{figure}

\paragraph{Experimental Results} 
Figure~\ref{fig:4.22} shows the results. 
Overall, the models trained with automatically generated datasets consistently outperformed unsupervised methods.
Specifically, the 5-shot$\times$4 setting achieved the highest score of 82.71.
Comparing the performance of PromptEOL without fine-tuning and PromptEOL fine-tuned with the automatically generated dataset using zero-shot learning, the fine-tuned model consistently outperformed.
This indicates that fine-tuning with the generated NLI dataset is effective when no manually created examples are available.
Moreover, our models outperformed PromptRoBERTa, indicating that our model achieved the best performance without using large manually annotated datasets.

Compared to the model fine-tuned with the manual dataset, the performance of the 5-shot$\times$4 setting was 2--3 points lower.
This indicates that there is still a gap between the 5-shot$\times$4 dataset and the manual dataset, suggesting room for improvement.
Despite this gap, there was an approximately 10-point performance improvement compared to the model without fine-tuning, confirming the effectiveness of the automatically generated dataset.
We provide the detailed experimental results in Appendix~\ref{appendix:Detailed_STS_Scores}.

\section{Conclusion and Future Work}
In this study, we explore optimal ways to leverage few-shot examples when using LLMs to generate NLI datasets for sentence embedding learning.
Through experiments, we found that the performance could be enhanced by dividing the few-shot examples, as seen with the 5-shot$\times$4 setting, since it improves dataset diversity.
Furthermore, models trained with automatically generated NLI datasets outperformed existing unsupervised methods.



In future work, we will explore more sophisticated ways to generate a diverse and high-quality dataset.
For example, instead of just dividing few-shot examples, a set of various overlapping few-shot examples could be generated and used in few-shot learning.
It is also future work to apply our data generation procedure, which generates data by dividing few-shot examples, to data generation other than NLI datasets for sentence embedding learning.  


\section*{Limitations}
There are three major limitations in this study.
Firstly, we only conducted experiments using LLaMA-2-7B as the LLM for both the automatic generation of the NLI dataset and the generation of sentence embeddings.
It is known that the quality of generated sentences improves as the number of parameters in the LLM increases.
In this study as well, it may be possible to obtain higher quality NLI datasets and sentence embedding models by using a model larger than LLaMA-2-7B.
Since this method is expected to be applicable to many LLMs without depending on a specific LLM, to demonstrate the model-independent usefulness of our observations, we need to conduct experiments using various LLMs, such as the GPT series and OPT~\cite{OPT}.

Secondly, we followed the previous research, PromptEOL, and conducted evaluations using SentEval.
However, it is not enough to comprehensively assess the quality of sentence embeddings to evaluate only with the STS tasks and SentEval.
It is necessary to use various benchmarks, such as MTEB~\cite{MTEB}, which evaluate sentence embeddings from multiple perspectives.

Thirdly, the experiments were conducted only in English.
It is potentially applicable to many languages to generate datasets automatically because it does not require large, manually-annotated datasets, but our experiments were conducted only in English.
To demonstrate the usefulness of our observation for multiple languages and improve cross-lingual/multi-lingual sentence embeddings, it is helpful to conduct experiments in languages other than English.

\section*{Acknowledgements}
This work was partly supported by JSPS KAKENHI Grant Number 24H007271.

\bibliography{anthology,custom}

\appendix
\begin{figure}[ht!]
\centering
\includegraphics[width=\linewidth]{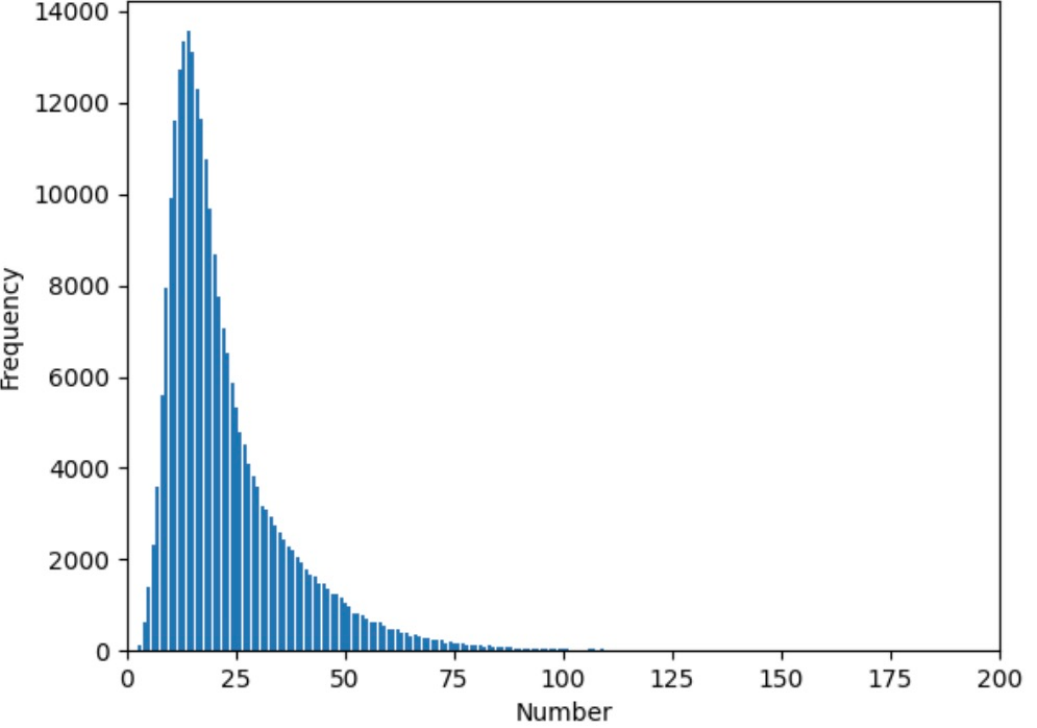}
\caption{Frequency distribution of token counts in the manual NLI dataset}
\label{fig:distribution}
\end{figure}

\begin{table*}[ht!]
\small
\centering
\begin{tabular}{lll}
\hline
\tabH Assigned label & Premise and generated hypothesis sentence & Predicted label\\
\hline
\tabH Premise & It concluded in July 2019. & -\\
Entailment & It was completed in July 2019. & \textbf{Entailment}\\
Contradiction & It did not conclude in July 2019. & \textbf{Contradiction}\\
\hline
\tabH Premise & He spent several months in prison. & -\\ 
Entailment & He was unable to pay his bills while in prison. & Neutral\\
Contradiction & He was not in prison for any amount of time. &\textbf{Contradiction}\\
\hline
\tabH Premise & Her last public performance was in 1954. & -\\
Entailment & She has not performed in public since 1954.&\textbf{Entailment}\\
Contradiction & Her last public performance was last week.&\textbf{Contradiction}\\
\hline
\tabH Premise & It grows on a many different soils. & -\\
Entailment & It grows on any soil that is suitable for the growth of other crops. & Neutral\\
Contradiction & It does not grow on soil with high pH levels. & Neutral\\
\hline
\end{tabular}
\caption{
Examples of the NLI dataset generated automatically with zero-shot learning.
The predicted labels matched the assigned label are shown in \textbf{bold}.
}
\label{tab:0_nli_example}
\end{table*}

\begin{table*}[ht!]
\small
\centering
\begin{tabular}{lll}
\hline
\tabH Assigned label & Premise and generated hypothesis sentence & Predicted label\\
\hline
\tabH Premise & It concluded in July 2019. & - \\
Entailment & July 2019 occurred. &\textbf{Entailment}\\
Contradiction & It began in January 2020.&\textbf{Contradiction}\\
\hline
\tabH Premise & He spent several months in prison. & - \\
Entailment & He was in prison for several months. &\textbf{Entailment}\\
Contradiction & He was never in prison. &\textbf{Contradiction}\\
\hline
\tabH Premise & Her last public performance was in 1954. & -\\
Entailment & She performed in 1954.&\textbf{Entailment}\\
Contradiction & She is still actively touring and performing.&\textbf{Contradiction}\\
\hline
\tabH Premise & It grows on a many different soils.& - \\
Entailment & The plant grows on various soils.&\textbf{Entailment}\\
Contradiction & It only grows on sandy soils. &\textbf{Contradiction}\\
\hline
\end{tabular}
\caption{
Examples of the NLI dataset generated automatically with 20-shot learning.
The predicted labels matched the assigned label are shown in \textbf{bold}.
}
\label{tab:20_nli_example}
\end{table*}

\section{Frequency Distribution of Token Counts in the Manual NLI Dataset}
\label{appendix:distribution}
Figure~\ref{fig:distribution} shows the frequency distribution of the token counts in the manual NLI dataset. 
The counts for most sentences are distributed between 1 and 100, with about 83.0\% of them having counts between 4 and 32.
Accordingly, we used sentences within that range in this work.

\section{Examples of Automatically Generated NLI Datasets}
\label{appendix:NLI_examples}
Tables~\ref{tab:0_nli_example} and \ref{tab:20_nli_example} provide examples of NLI data that were automatically generated with 0-shot and 20-shot learning, respectively, for the same premise sentences.

We observed that some sentences generated in the zero-shot setting are predicted as neutral, but sentences close to entailment and contradiction can also be generated.
By shifting the period or using negation, diverse entailment and contradiction sentences can be created.
The sentences generated with 20-shot learning tended to strongly refer to the premise sentence, indicating higher precision in generating both entailment and contradiction sentences.
Additionally, these sentences tended to be shorter than those generated with 0-shot learning.

\section{Evaluation of Transfer Tasks}
\label{appendix:transfer_tasks}
To evaluate the effectiveness of the generated sentence embeddings in transfer tasks, we conducted evaluations with transfer tasks from SentEval~\cite{SentEval}.
These tasks use sentence embeddings as input and train a linear classifier.
Specifically, the embeddings generated from each sentence are used as features to train linear classifiers, such as logistic regression.
The classification task performance is assessed with the trained classifier, and the accuracy and other related metrics are measured to quantitatively evaluate the effectiveness of the sentence representations.

Table~\ref{tab:transfer_256000} summarizes the results.
As reported in \citet{PromptEOL}, we could not confirm any performance improvement on the transfer tasks via fine-tuning; however, the scores for both the zero-shot and few-shot settings were comparable to those with training on the manual NLI dataset.
Tables~\ref{tab:transfer_4.21}~and~\ref{tab:transfer_4.22} show detailed scores for each experiment.
In the proposed method, it is confirmed that the score is low and unstable when the data size is small, but it stabilizes as the data size increases.

\section{Detailed STS Scores}
\label{appendix:Detailed_STS_Scores}
Table~\ref{tab:sts_256000} shows the performances of the STS tasks for each model with 256,000 examples.
Additionally, Tables~\ref{tab:experiment4.21}~and~\ref{tab:experiment4.22} show detailed performances of the STS tasks for each experiment.
It is evident that good sentence embedding models have been created without any extreme highs or lows for any dataset. 
Furthermore, the 1-shot setting tends to have a larger variance, while the variance tends to decrease as the number of shots increases. 
This confirms the validity of increasing the number of trials as the number of shots increases.
\begin{table*}[ht!]
\small
\centering
\begin{tabular}{lccccccccc}
\hline
\tabH Model   & MR & CR & SUBJ & MPQA & SST & TREC & MRPC & Avg.  \\
\hline
\multicolumn{9}{c}{\tabH \footnotesize{\textbf{Without fine-tuning (base model: LLaMA-2-7B)}}}\\
\tabH PromptEOL  & 90.53 & 92.45 & 96.22 & 91.24 & 95.39 & 96.20  & 74.96 & 91.00$_{\pm 0.000}$\\
\hline
\multicolumn{9}{c}{\tabH \footnotesize{\textbf{Fine-tuning on unsupervised dataset}}}\\
\tabH PromptRoBERTa  &82.88	&88.14	&94.13	&87.22	&87.97	&88.60	&74.63	&86.22$_{\pm 0.159}$\\
\hline
\multicolumn{9}{c}{\tabH \footnotesize{\textbf{Fine-tuning on automatically generated dataset (base model: LLaMA-2-7B)}}}\\
\tabH PromptEOL (0-shot)   &90.00	&92.58	&95.23	&90.56	&94.07	&94.00	&73.39	&89.97$_{\pm 0.511}$ \\
PromptEOL (1-shot)  &89.93	&92.63	&95.28	&90.62	&94.32	&94.76	&72.17	&89.96$_{\pm 0.248}$ \\
PromptEOL (1-shot$\times$5)    &90.38	&92.75	&95.49	&90.61	&94.53	&95.28	&72.79	&90.26$_{\pm 0.312}$ \\
PromptEOL (5-shot)   &89.63	&92.52	&94.74	&90.68	&93.84	&95.16	&73.77	&90.05$_{\pm 0.086}$ \\
PromptEOL (5-shot$\times$4)    &89.63	&92.52	&94.74	&90.68	&93.84	&95.16	&73.77	&90.05$_{\pm 0.293}$ \\
PromptEOL (20-shot)  &89.33	&92.76	&94.71	&91.19	&93.66	&93.65	&74.20	&89.93$_{\pm 0.240}$ \\
\hline
\multicolumn{9}{c}{\tabH \footnotesize{\textbf{Fine-tuning on manual dataset (base model: LLaMA-2-7B)}}}\\
\tabH PromptEOL   &89.94	&93.22	&96.05	&90.83	&94.89	&95.40	&74.26	&90.65$_{\pm 0.227}$\\
\hline
\end{tabular}
\caption{
Transfer task results of different sentence embedding models (measured as accuracy). 
256,000 sentence pairs were used for fine-tuning the model.
The average performance (Avg.) is provided along with the respective standard deviation.
}
\label{tab:transfer_256000}
\end{table*}

\begin{table*}[t!]
\small
\centering
\begin{tabular}{cccccccccc}
\hline
\tabH Dataset size & Setting  & MR & CR & SUBJ & MPQA & SST & TREC & MRPC & Avg. \\
\hline

\tabH \ \ \ \ 4000 & \multirow{7}{*}{0-shot}  &90.08	&36.24	&54.58	&\textbf{90.87}	&\textbf{94.47}	&\verb| |1.80	&33.51	&57.36$_{\pm 0.045}$ \\
\ \ \ \ 8000 &  &90.26	&92.25	&95.65	&90.27	&93.57	&\textbf{94.73}	&71.92	&89.81$_{\pm 0.172}$ \\
\ \ 16000  &  &90.24	&92.30	&\textbf{95.73}	&90.44	&79.33	&94.60	&59.56	&86.03$_{\pm 5.630}$ \\
\ \ 32000   & &76.72	&73.47	&80.37	&90.36	&79.39	&63.53	&60.21	&74.87$_{\pm 21.39}$ \\
\ \ 64000  &  &\textbf{90.46}	&\textbf{92.64}	&95.71	&90.67	&94.23	&\textbf{94.73}	&72.89	&\textbf{90.19}$_{\pm 0.065}$ \\
128000 &   &89.83	&92.49	&95.61	&90.34	&94.40	&94.33	&\textbf{73.74}	&90.10$_{\pm 0.143}$ \\
256000 &   &90.00	&92.58	&95.23	&90.56	&94.07	&94.00	&73.39	&89.97$_{\pm 0.511}$ \\
\hline
\tabH \ \ \ \ 4000 &\multirow{7}{*}{1-shot}  &85.75	&48.61	&55.01	&88.57	&58.99	&56.98	&33.51	&61.06$_{\pm 12.59}$ \\
\ \ \ \ 8000 &  &89.84	&92.50	&95.14	&90.44	&93.85	&94.74	&65.68	&88.88$_{\pm 2.359}$ \\
\ \ 16000 &  &89.78	&92.62	&95.23	&90.63	&93.78	&94.48	&\textbf{72.89}	&89.92$_{\pm 0.500}$ \\
\ \ 32000 &  &89.66	&87.21	&91.01	&90.59	&89.67	&\textbf{94.88}	&68.63	&87.38$_{\pm 6.002}$ \\
\ \ 64000 &  &\textbf{90.23}	&87.18	&90.22	&\textbf{91.11}	&89.73	&91.86	&70.49	&78.27$_{\pm 8.334}$ \\
128000 &  &90.09	&87.43	&82.26	&90.58	&94.17	&94.36	&53.26	&84.59$_{\pm 6.580}$ \\
256000 & &89.93	&\textbf{92.63}	&\textbf{95.28}	&90.62	&\textbf{94.32}	&94.76	&72.17	&\textbf{89.96}$_{\pm 0.248}$ \\
\hline
\tabH \ \ \ \ 4000 &\multirow{7}{*}{1-shot$\times$5}  &\textbf{90.46}	&48.67	&50.00	&\textbf{90.63}	&67.98	&57.84	&33.51	&62.73$_{\pm 9.295}$ \\
\ \ \ \ 8000 & &90.29	&92.23	&79.04	&90.50	&85.45	&94.84	&72.98	&86.48$_{\pm 4.861}$ \\
\ \ 16000 &  &90.43	&92.30&95.46	&90.42	&94.29	&95.08	&\textbf{74.15}	&\textbf{90.30}$_{\pm 0.221}$ \\
\ \ 32000 & &90.16	&92.58	&95.03	&90.56	&93.97	&94.96	&74.11	&90.20$_{\pm 0.055}$ \\
\ \ 64000 &  &90.20	&92.39	&95.45	&90.62	&94.08	&94.80	&73.14	&90.10$_{\pm 0.292}$ \\
128000 &  &90.12	&92.57	&95.46	&90.49	&\textbf{94.71}	&\textbf{95.48}	&73.19	&90.29$_{\pm 0.265}$ \\
256000 &  &90.38	&\textbf{92.75}	&\textbf{95.49}	&90.61	&94.53	&95.28	&72.79	&90.26$_{\pm 0.312}$ \\
\hline
\tabH \ \ \ \ 4000 &\multirow{7}{*}{5-shot} &81.68	&81.45	&68.82	&86.71	&85.06	&57.76	&58.04	&74.22$_{\pm 18.80}$ \\
\ \ \ \ 8000 &  &81.95	&70.21	&76.97	&90.67	&76.69	&57.96	&58.44	&73.27$_{\pm 21.13}$\\
\ \ 16000 &   &89.83	&92.73	&94.94	&90.76	&\textbf{94.00}	&94.96	&74.06	&\textbf{90.19}$_{\pm 0.571}$ \\
\ \ 32000 & &\textbf{89.95}	&92.80	&94.98	&\textbf{90.95}	&93.81	&94.80	&73.84	&90.16$_{\pm 0.354}$ \\
\ \ 64000 &   &89.44	&92.72	&94.82	&90.74	&93.77	&95.00	&\textbf{74.25}	&90.10$_{\pm 0.289}$\\
128000 &   &89.78	&\textbf{92.85}	&\textbf{95.04}	&90.72	&93.77	&94.72	&73.43	&90.04$_{\pm 0.310}$\\
256000 &   &89.63	&92.52	&94.74	&90.68	&93.84	&\textbf{95.16}	&73.77	&90.05$_{\pm 0.293}$ \\
\hline
\tabH \ \ \ \ 4000 &\multirow{7}{*}{5-shot$\times$4}  &79.60	&50.30	&50.00	&85.72	&61.22	&25.20	&33.51	&55.08$_{\pm 13.70}$ \\
\ \ \ \ 8000 &    &\textbf{89.86}	&80.20	&72.67	&90.71	&83.77	&94.60	&43.84	&79.38$_{\pm 9.888}$ \\
\ \ 16000 &    &89.52	&66.02	&73.66	&90.79	&82.84	&71.85	&54.02	&75.53$_{\pm 16.47}$ \\
\ \ 32000 &    &89.64	&80.08	&83.42	&90.71	&82.66	&\textbf{94.65}	&63.57	&83.53$_{\pm 11.15}$ \\
\ \ 64000 &  &89.39	&\textbf{93.10}	&94.75	&\textbf{91.01}	&93.37	&94.25	&\textbf{73.57}	&\textbf{89.92}$_{\pm 0.303}$ \\
128000 &    &89.74	&92.84	&\textbf{95.00}	&90.89	&93.93	&93.65	&63.49	&88.51$_{\pm 2.530}$ \\
256000 &   &89.64	&92.76	&94.84	&90.62	&\textbf{94.13}	&94.10	&73.07	&89.88$_{\pm 0.264}$ \\
\hline
\tabH \ \ \ \ 4000 &\multirow{7}{*}{20-shot}  &79.61	&36.24	&50.00	&80.78	&50.08	&\verb| |1.80	&33.51	&47.43$_{\pm 3.983}$ \\
\ \ \ \ 8000 & &70.49	&36.24	&50.00	&75.84	&50.08	&24.90	&33.51	&48.73$_{\pm 9.173}$ \\
\ \ 16000 & &\textbf{89.68}	&78.65	&72.32	&90.54	&82.22	&70.60	&54.20	&76.89$_{\pm 18.61}$ \\
\ \ 32000 &  &89.19	&\textbf{92.86}	&94.60	&90.51	&93.25	&\textbf{94.65}	&74.12	&89.88$_{\pm 0.070}$ \\
\ \ 64000 &  &89.62	&92.78	&\textbf{94.89}	&91.00	&93.84	&94.15	&\textbf{74.47}	&\textbf{90.11}$_{\pm 0.205}$ \\
128000 & &88.89	&92.82	&94.67	&90.92	&92.92	&94.30	&74.21	&89.82$_{\pm 0.146}$ \\
256000 & &89.33	&92.76	&94.71	&\textbf{91.19}	&\textbf{93.66}	&93.65	&74.20	&89.93$_{\pm 0.240}$ \\
\hline
\end{tabular}
\caption{
The results of PromptEOL-LLaMA-2-7B fine-tuned with the automatically generated dataset (measured as accuracy).
The average performance (Avg.) is provided along with the respective standard deviation.
}
\label{tab:transfer_4.21}
\end{table*}

\begin{table*}[t!]
\small
\centering
\begin{tabular}{lccccccccc}
\hline
\tabH Model & Dataset size & MR & CR & SUBJ & MPQA & SST & TREC & MRPC & Avg. \\
\hline
\multicolumn{10}{c}{\tabH \footnotesize{\textbf{Without fine-tuning (base model: LLaMA-2-7B)}}}\\
\tabH PromptEOL  &- & 59.91 & 78.86 & 68.74 & 75.71 & 73.39 & 73.48  & 71.26 & 71.62$_{\pm 0.000}$\\
\hline
\multicolumn{10}{c}{\tabH \footnotesize{\textbf{Fine-tuning on unsupervised dataset}}}\\
\tabH \multirow{7}{*}{PromptRoBERTa} &\ \ \ \ 4000  &\textbf{83.90}	&\textbf{88.78}	&\textbf{95.31}	&86.72	&\textbf{89.16}	&\textbf{93.73}	&74.07	&\textbf{87.38}$_{\pm 0.076}$\\
 &\ \ \ \ 8000 &83.58	&88.54	&95.31	&86.53	&88.65	&91.67	&74.36	&86.95$_{\pm 0.081}$\\
 &\ \ 16000 &83.19	&87.41	&94.93	&86.66	&88.17	&90.40	&73.43	&86.32$_{\pm 0.017}$\\
 &\ \ 32000 &83.06	&87.58	&94.64	&86.84	&87.99	&88.47	&73.35	&85.99$_{\pm 0.080}$\\
 &\ \ 64000 &82.96	&87.72	&94.42	&87.01	&87.66	&88.80	&74.14	&86.10$_{\pm 0.033}$\\
 &128000 &82.58	&87.73	&94.18	&86.94	&87.66	&88.47	&74.13	&85.95$_{\pm 0.186}$\\
 &256000  &82.88	&88.14	&94.13	&\textbf{87.22}	&87.97	&88.60	&\textbf{74.63}	&86.22$_{\pm 0.159}$\\
\hline
\multicolumn{10}{c}{\tabH \footnotesize{\textbf{Fine-tuning on automatically generated dataset (base model: LLaMA-2-7B)}}}\\
\tabH \multirow{7}{*}{PromptEOL (0-shot)} 
&\ \ \ \ 4000  &90.08	&36.24	&54.58	&\textbf{90.87}	&\textbf{94.47}	&\verb| |1.80	&33.51	&57.36$_{\pm 0.045}$ \\
&\ \ \ \ 8000  &90.26	&92.25	&95.65	&90.27	&93.57	&\textbf{94.73}	&71.92	&89.81$_{\pm 0.172}$ \\
&\ \ 16000  &90.24	&92.30	&\textbf{95.73}	&90.44	&79.33	&94.60	&59.56	&86.03$_{\pm 5.630}$ \\
&\ \ 32000 &76.72	&73.47	&80.37	&90.36	&79.39	&63.53	&60.21	&74.87$_{\pm 21.39}$ \\
&\ \ 64000 &\textbf{90.46}	&92.64	&95.71	&90.67	&94.23	&\textbf{94.73}	&72.89	&\textbf{90.19}$_{\pm 0.065}$ \\
&128000   &89.83	&92.49	&95.61	&90.34	&94.40	&94.33	&\textbf{73.74}	&90.10$_{\pm 0.143}$ \\
&256000  &90.00	&\textbf{92.58}	&95.23	&90.56	&94.07	&94.00	&73.39	&89.97$_{\pm 0.511}$ \\
 \hdashline
\tabH \multirow{7}{*}{PromptEOL (5-shot$\times$4)}& \ \ \ \ 4000  &79.60	&50.30	&50.00	&85.72	&61.22	&25.20	&33.51	&55.08$_{\pm 13.70}$ \\
&\ \ \ \ 8000    &\textbf{89.86}	&80.20	&72.67	&90.71	&83.77	&94.60	&43.84	&79.38$_{\pm 9.888}$ \\
&\ \ 16000    &89.52	&66.02	&73.66	&90.79	&82.84	&71.85	&54.02	&75.53$_{\pm 16.47}$ \\
&\ \ 32000    &89.64	&80.08	&83.42	&90.71	&82.66	&\textbf{94.65}	&63.57	&83.53$_{\pm 11.15}$ \\
&\ \ 64000  &89.39	&\textbf{93.10}	&94.75	&\textbf{91.01}	&93.37	&94.25	&\textbf{73.57}	&\textbf{89.92}$_{\pm 0.303}$ \\
&128000    &89.74	&92.84	&\textbf{95.00}	&90.89	&93.93	&93.65	&63.49	&88.51$_{\pm 2.530}$ \\
&256000   &89.64	&92.76	&94.84	&90.62	&\textbf{94.13}	&94.10	&73.07	&89.88$_{\pm 0.264}$ \\
\hline
\multicolumn{10}{c}{\tabH \footnotesize{\textbf{Fine-tuning on manually annotated dataset (base model: LLaMA-2-7B)}}}\\
\tabH \multirow{7}{*}{PromptEOL}&\ \ \ \ 4000  &88.06	&92.14	&66.25	&90.53	&92.39	&93.27	&73.93	&85.23$_{\pm 2.435}$\\
 &\ \ \ \ 8000 &88.18	&92.82	&94.90	&90.39	&93.14	&94.00	&\textbf{74.80}	&89.75$_{\pm 0.215}$\\
 &\ \ 16000 &88.81	&92.93	&81.54	&90.34	&79.11	&92.93	&73.51	&85.59$_{\pm 5.488}$\\
 &\ \ 32000 &89.68	&93.13	&95.34	&90.20	&79.74	&95.00	&73.39	&88.07$_{\pm 3.070}$\\
 &\ \ 64000 &89.78	&\textbf{93.30}	&96.02	&90.54	&\textbf{94.89}	&95.33	&73.80	&90.52$_{\pm 0.041}$\\
 &128000 &\textbf{90.12}	&93.07	&\textbf{96.07}	&90.82	&94.33	&95.33	&74.30	&90.57$_{\pm 0.131}$\\
 &256000  &89.94	&93.22	&96.05	&\textbf{90.83}	&\textbf{94.89}	&\textbf{95.40}	&74.26	&\textbf{90.65}$_{\pm 0.227}$\\
\hline
\end{tabular}
\caption{
Transfer task results of different sentence embedding models (measured as accuracy).
The average performance (Avg.) is provided along with the respective standard deviation.
}
\label{tab:transfer_4.22}
\end{table*}


\begin{table*}[ht!]
\small
\centering
\begin{tabular}{lccccccccc}
\hline
\tabH Model & STS12  & STS13 & STS14 & STS15 & STS16 & STS-B & SICK-R & Avg.  \\
\hline
\multicolumn{9}{c}{\tabH \footnotesize{\textbf{Without fine-tuning (base model: LLaMA-2-7B)}}}\\
\tabH PromptEOL & 59.91 & 78.86 & 68.74 & 75.71 & 73.39 & 73.48  & 71.26 & 71.62$_{\pm 0.000}$\\
\hline
\multicolumn{9}{c}{\tabH \footnotesize{\textbf{Fine-tuning on unsupervised dataset}}}\\
\tabH PromptRoBERTa  &73.64	&84.97	&77.44	&85.11	&81.61	&82.12	&69.09	&79.14$_{\pm 0.175}$\\
\hline
\multicolumn{9}{c}{\tabH \footnotesize{\textbf{Fine-tuning on automatically generated dataset (base model: LLaMA-2-7B)}}}\\
\tabH PromptEOL (0-shot)   &71.76	&86.47	&80.53	&83.26	&83.75	&82.45	&71.95	&80.02$_{\pm 0.485}$ \\
PromptEOL (1-shot) &73.30	&87.61	&81.52	&85.35	&83.85	&83.63	&76.86	&81.73$_{\pm 1.140}$ \\
PromptEOL  (1-shot$\times$5)  &73.27	&87.90	&81.74	&85.72	&84.11	&84.66	&76.45	&81.98$_{\pm 0.837}$ \\
PromptEOL  (5-shot)   &73.72	&87.75	&81.94	&85.71	&83.85	&84.49	&75.72	&81.88$_{\pm 0.846}$\\
PromptEOL  (5-shot$\times$4)   &74.16	&87.75	&82.65	&85.95	&84.97	&85.26	&76.44	&82.45$_{\pm 0.385}$  \\
PromptEOL  (20-shot) &74.24	&87.39	&82.55	&85.49	&84.36	&85.24	&74.20	&81.92$_{\pm 0.183}$ \\
\hline
\multicolumn{9}{c}{\tabH \footnotesize{\textbf{Fine-tuning on manual dataset (base model: LLaMA-2-7B)}}}\\
\tabH PromptEOL   &78.75	&89.99	&84.98	&88.82	&86.27	&88.37	&82.44	&85.66$_{\pm 0.101}$\\
\hline
\end{tabular}
\caption{
Spearman's rank correlation coefficient between the cosine similarity of the sentence embeddings and the human ratings.
All values in the table are multiplied by 100.
256,000 sentence pairs were used for fine-tuning the model.
The average performance (Avg.) is provided along with the respective standard deviation.
}
\label{tab:sts_256000}
\end{table*}

\begin{table*}[t!]
\small
\centering
\begin{tabular}{ccccccccccc}
\hline
\tabH Dataset size & few-shot & STS12  & STS13 & STS14 & STS15 & STS16 & STS-B & SICK-R & Avg. \\
\hline
\tabH \ \ \ \ 4000 &\multirow{7}{*}{0-shot}   &65.68	&83.92	&76.18	&80.00	&79.95	&78.82	&\textbf{76.80}	&77.34$_{\pm 0.185}$ \\
\ \ \ \ 8000 &   &68.80	&85.60	&78.42	&81.51	&81.79	&81.26	&73.16	&78.65$_{\pm 0.159}$ \\
\ \ 16000 &   &69.37	&85.62	&77.97	&81.56	&81.99	&81.76	&74.37	&78.95$_{\pm 0.362}$ \\
\ \ 32000 &   &71.59	&85.93	&78.40	&82.34	&82.24	&81.51	&74.44	&79.49$_{\pm 0.553}$ \\
\ \ 64000 &   &71.21	&86.09	&80.28	&\textbf{83.60}	&83.21	&81.43	&73.19	&79.86$_{\pm 0.137}$ \\
128000 &   &70.84	&86.40	&80.15	&83.12	&82.44	&82.29	&73.71	&79.85$_{\pm 0.310}$ \\
256000 &   &\textbf{71.76}	&\textbf{86.47}	&\textbf{80.53}	&83.26	&\textbf{83.75}	&\textbf{82.45}	&71.95	&\textbf{80.02}$_{\pm 0.485}$ \\
\hline
\tabH \ \ \ \ 4000 &\multirow{7}{*}{1-shot}   &70.14	&85.97	&78.94	&82.05	&82.42	&82.30	&\textbf{77.38}	&79.89$_{\pm 1.038}$ \\
\ \ \ \ 8000 &  &73.03	&\textbf{87.82}	&81.52	&85.47	&83.79	&\textbf{84.57}	&77.13	&\textbf{81.90}$_{\pm 0.764}$ \\
\ \ 16000 &  &72.92	&87.63	&81.32	&85.29	&83.67	&84.28	&76.96	&81.73$_{\pm 0.959}$ \\
\ \ 32000 &  &72.82	&87.34	&81.05	&84.99	&83.38	&83.71	&76.76	&81.44$_{\pm 2.185}$ \\
\ \ 64000 &  &\textbf{73.30}	&87.61	&81.52	&85.35	&\textbf{83.85}	&83.63	&76.86	&81.73$_{\pm 1.608}$ \\
128000 &  &73.01	&87.59	&\textbf{81.55}	&\textbf{85.51}	&83.77	&84.42	&76.76	&81.80$_{\pm 0.785}$ \\
256000 & &\textbf{73.30}	&87.61	&81.52	&85.35	&\textbf{83.85}	&83.63	&76.86	&81.73$_{\pm 1.140}$ \\
\hline
\tabH \ \ \ \ 4000 &\multirow{7}{*}{1-shot$\times$5}   &69.67	&86.03	&78.96	&82.63	&82.74	&82.28	&76.72	&79.86$_{\pm 0.846}$ \\
\ \ \ \ 8000 &  &73.05	&87.43	&81.40	&84.95	&83.60	&84.30	&77.15	&81.70$_{\pm 0.920}$ \\
\ \ 16000 &  &72.94	&87.79	&81.57	&85.03	&83.70	&\textbf{84.56}	&\textbf{77.26}	&81.83$_{\pm 0.544}$ \\
\ \ 32000 & &72.83	&87.81	&81.26	&84.88	&83.54	&83.97	&76.81	&81.59$_{\pm 0.198}$ \\
\ \ 64000 &  &\textbf{74.28}	&\textbf{87.85}	&\textbf{81.92}	&85.53	&\textbf{84.13}	&84.21	&75.87	&81.97$_{\pm 0.370}$ \\
128000 &  &72.60	&87.50	&81.32	&85.35	&83.45	&84.18	&75.72	&81.44$_{\pm 0.379}$ \\
256000 &  &73.27	&87.90	&81.74	&\textbf{85.72}	&84.11	&84.66	&76.45	&\textbf{81.98}$_{\pm 0.837}$ \\
\hline
\tabH \ \ \ \ 4000 &\multirow{7}{*}{5-shot}  &70.40	&85.99	&79.35	&82.48	&82.61	&82.15	&76.11	&79.87$_{\pm 3.850}$ \\
\ \ \ \ 8000 &   &72.58	&87.23	&80.92	&84.63	&83.47	&84.22	&76.55	&81.37$_{\pm 1.962}$ \\
\ \ 16000 &   &73.34	&87.52	&81.52	&85.53	&83.57	&84.72	&\textbf{76.93}	&81.88$_{\pm 1.081}$ \\
\ \ 32000 &  &\textbf{73.81}	&87.33	&\textbf{81.78}	&85.27	&83.67	&\textbf{84.86}	&76.01	&81.82$_{\pm 0.947}$ \\
\ \ 64000 &   &73.50	&87.75	&81.67	&\textbf{85.76}	&83.71	&84.69	&76.67	&\textbf{81.97}$_{\pm 1.109}$ \\
128000 &   &73.68	&87.52	&81.65	&85.25	&83.57	&84.85	&74.71	&81.60$_{\pm 1.111}$ \\
256000 &   &73.72	&\textbf{87.75}	&81.94	&85.71	&\textbf{83.85}	&84.49	&75.72	&81.88$_{\pm 0.846}$ \\
\hline
\tabH \ \ \ \ 4000 & \multirow{7}{*}{5-shot$\times$4} &71.82	&87.20	&80.36	&83.39	&83.63	&83.46	&74.88	&80.68$_{\pm 0.686}$ \\
\ \ \ \ 8000 &    &73.10	&87.83	&82.08	&84.98	&84.21	&84.89	&\textbf{76.70}	&81.97$_{\pm 0.207}$ \\
\ \ 16000 &    &73.63	&\textbf{88.01}	&82.38	&85.55	&84.05	&85.25	&75.87	&82.10$_{\pm 0.416}$ \\
\ \ 32000 &    &74.67	&87.92	&82.02	&85.68	&84.52	&85.43	&75.72	&82.28$_{\pm 0.445}$ \\
\ \ 64000 &   &\textbf{74.88}	&87.93	&\textbf{82.83}	&\textbf{86.23}	&84.68	&\textbf{86.00}	&76.40	&\textbf{82.71}$_{\pm 0.322}$ \\
128000 &  &74.11	&87.55	&82.00	&85.51	&84.22	&85.30	&75.78	&82.06$_{\pm 0.209}$ \\
256000 &  &74.16	&87.75	&82.65	&85.95	&\textbf{84.97}	&85.26	&76.44	&82.45$_{\pm 0.385}$ \\
\hline
\tabH \ \ \ \ 4000 &\multirow{7}{*}{20-shot}  &71.48	&86.75	&80.11	&82.72	&83.08	&83.67	&74.44	&80.32$_{\pm 0.472}$ \\
\ \ \ \ 8000 & &72.55	&87.38	&81.46	&83.75	&83.14	&84.62	&76.01	&81.27$_{\pm 0.093}$ \\
\ \ 16000 & &73.17	&87.10	&81.54	&84.79	&83.47	&84.97	&75.50	&81.50$_{\pm 0.575}$ \\
\ \ 32000 &  &74.11	&87.67	&82.14	&\textbf{85.59}	&84.22	&\textbf{85.56}	&\textbf{75.83}	&\textbf{82.16}$_{\pm 0.315}$ \\
\ \ 64000 &  &73.73	&87.24	&81.69	&84.78	&83.76	&85.00	&74.83	&81.58$_{\pm 0.182}$ \\
128000 &  &74.15	&\textbf{87.44}	&82.36	&85.16	&83.91	&85.30	&74.38	&81.81$_{\pm 0.558}$ \\
256000 & &\textbf{74.24}	&87.39	&\textbf{82.55}	&85.49	&\textbf{84.36}	&85.24	&74.20	&81.92$_{\pm 0.183}$ \\
\hline
\end{tabular}
\caption{
The detailed results of the experiments conducted in Section~\ref{sec:sts}.
Spearman's rank correlation coefficient between the cosine similarity of the sentence embeddings of PromptEOL-LLaMA-2-7B fine-tuned with an automatically generated dataset with few-shot and the human evaluation.
All values in the table are multiplied by 100.
The average performance (Avg.) is provided along with the respective standard deviation.
}
\label{tab:experiment4.21}
\end{table*}

\begin{table*}[t!]
\small
\centering
\tabcolsep 4.5pt
\begin{tabular}{lccccccccc}
\hline
\tabH Model &Dataset size & STS12  & STS13 & STS14 & STS15 & STS16 & STS-B & SICK-R & Avg. \\
\hline
\multicolumn{10}{c}{\tabH \footnotesize{\textbf{Without fine-tuning (base model: LLaMA-2-7B)}}}\\
PromptEOL  &- & 59.91 & 78.86 & 68.74 & 75.71 & 73.39 & 73.48  & 71.26 & 71.62$_{\pm 0.000}$\\
\hline
\multicolumn{10}{c}{\tabH \footnotesize{\textbf{Fine-tuning on unsupervised dataset}}}\\
\tabH \multirow{7}{*}{PromptRoBERTa} &\ \ \ \ 4000  &62.74	&80.97	&70.30	&80.75	&76.78	&77.49	&\textbf{71.24}	&74.33$_{\pm 0.078}$\\
 &\ \ \ \ 8000 &61.86	&79.56	&69.67	&81.05	&75.52	&76.15	&70.78	&73.51$_{\pm 0.156}$\\
 &\ \ 16000 &66.67	&81.03	&72.17	&82.87	&77.67	&78.90	&70.08	&75.63$_{\pm 0.264}$\\
 &\ \ 32000 &71.25	&84.01	&75.29	&84.43	&80.39	&81.00	&69.26	&77.95$_{\pm 0.064}$\\
 &\ \ 64000 &72.90	&84.59	&76.47	&84.92	&80.85	&81.60	&68.71	&78.58$_{\pm 0.111}$\\
 &128000 &72.98	&84.71	&76.80	&84.98	&80.96	&81.68	&68.77	&78.70$_{\pm 0.273}$\\
 &256000 &\textbf{73.64}	&\textbf{84.97}	&\textbf{77.44}	&\textbf{85.11}	&\textbf{81.61}	&\textbf{82.12}	&69.09	&\textbf{79.14}$_{\pm 0.175}$\\
\hline
\multicolumn{10}{c}{\footnotesize{\tabH \textbf{Fine-tuning on automatically generated dataset (base model: LLaMA-2-7B)}}}\\
\tabH \multirow{7}{*}{PromptEOL (0-shot)} &\ \ \ \ 4000  &65.68	&83.92	&76.18	&80.00	&79.95	&78.82	&\textbf{76.80}	&77.34$_{\pm 0.185}$ \\
 &\ \ \ \ 8000   &68.80	&85.60	&78.42	&81.51	&81.79	&81.26	&73.16	&78.65$_{\pm 0.159}$ \\
 &\ \ 16000  &69.37	&85.62	&77.97	&81.56	&81.99	&81.76	&74.37	&78.95$_{\pm 0.362}$ \\
 &\ \ 32000  &71.59	&85.93	&78.40	&82.34	&82.24	&81.51	&74.44	&79.49$_{\pm 0.553}$ \\
 &\ \ 64000  &71.21	&86.09	&80.28	&\textbf{83.60}	&83.21	&81.43	&73.19	&79.86$_{\pm 0.137}$ \\
 &128000&70.84	&86.40	&80.15	&83.12	&82.44	&82.29	&73.71	&79.85$_{\pm 0.310}$ \\
 &256000  &\textbf{71.76}	&\textbf{86.47}	&\textbf{80.53}	&83.26	&\textbf{83.75}	&\textbf{82.45}	&71.95	&\textbf{80.02}$_{\pm 0.485}$ \\
 \hdashline
\tabH \multirow{7}{*}{PromptEOL (5-shot$\times$4)} &\ \ \ \ 4000 &71.82	&87.20	&80.36	&83.39	&83.63	&83.46	&74.88	&80.68$_{\pm 0.686}$ \\
 &\ \ \ \ 8000  &73.10	&87.83	&82.08	&84.98	&84.21	&84.89	&\textbf{76.70}	&81.97$_{\pm 0.207}$ \\
 &\ \ 16000  &73.63	&\textbf{88.01}	&82.38	&85.55	&84.05	&85.25	&75.87	&82.10$_{\pm 0.416}$ \\
 &\ \ 32000  &\textbf{74.67}	&87.92	&82.02	&85.68	&84.52	&85.43	&75.72	&\textbf{82.28}$_{\pm 0.445}$ \\
 &\ \ 64000 &74.88	&87.93	&82.83	&\textbf{86.23}	&84.68	&\textbf{86.00}	&76.40	&82.71$_{\pm 0.322}$ \\
 &128000 &74.11	&87.55	&82.00	&85.51	&84.22	&85.30	&75.78	&82.06$_{\pm 0.209}$ \\
 &256000 &74.16	&87.75	&\textbf{82.65}	&85.95	&\textbf{84.97}	&85.26	&76.44	&82.45$_{\pm 0.385}$ \\
\hline
\multicolumn{10}{c}{\tabH \footnotesize{\textbf{Fine-tuning on manually annotated dataset (base model: LLaMA-2-7B)}}}\\
\tabH \multirow{7}{*}{PromptEOL} & \ \ \ \ 4000  &73.68	&87.41	&81.45	&86.06	&83.74	&86.18	&\textbf{82.85}	&83.05$_{\pm 0.423}$ \\
 &\ \ \ \ 8000  &74.93 &87.90	&82.61	&86.72	&84.67	&87.17	&82.83	&83.83$_{\pm 0.298}$ \\
 &\ \ 16000  &76.03	&87.99	&82.88	&87.24	&84.83	&87.21	&82.82	&84.14$_{\pm 0.504}$\\
 &\ \ 32000  &76.71	&88.26	&83.08	&87.22	&84.75	&87.52	&81.99	&84.22$_{\pm 1.270}$\\
 &\ \ 64000  &78.26	&89.64	&84.71	&\textbf{88.86}	&85.67	&88.18	&81.95	&85.32$_{\pm 0.117}$ \\
 &128000  &78.28	&89.89	&84.80	&\textbf{88.86}	&85.83	&88.35	&81.88	&85.41$_{\pm 0.172}$\\
 &256000  &\textbf{78.75}	&\textbf{89.99}	&\textbf{84.98}	&88.82	&\textbf{86.27}	&\textbf{88.37}	&82.44	&\textbf{85.66}$_{\pm 0.101}$\\
\hline
\end{tabular}
\caption{
The detailed results of the experiments conducted in Section~\ref{sec:4.3}. Spearman's rank correlation coefficient between the cosine similarity of the sentence embeddings and the human evaluation. All values in the table are multiplied by 100.
The average performance (Avg.) is provided along with the respective standard deviation.
}
\label{tab:experiment4.22}
\end{table*}

\end{document}